\pdfoutput=1

\documentclass[11pt]{article}

\usepackage[table]{xcolor}
\usepackage[]{acl}

\usepackage{times}
\usepackage{latexsym}

\usepackage[T1]{fontenc}

\usepackage[utf8]{inputenc}

\usepackage{microtype}

\usepackage{inconsolata}

\usepackage{graphicx}
\usepackage{amssymb}
\usepackage{pifont}

\usepackage{rotating}
\usepackage{amsmath}
\usepackage{multirow}
\usepackage{paralist}
\usepackage{listings}
\usepackage{makecell} 

\definecolor{lightpink}{rgb}{1.0, 0.71, 0.76} 
\definecolor{customred}{HTML}{fd7f6f}
\definecolor{customblue}{HTML}{7eb0d5}
\definecolor{customgreen}{HTML}{b2e061}
\definecolor{custompurple}{HTML}{bd7ebe}
\definecolor{customyellow}{HTML}{ffee65}

\title{Reference-based Metrics Disprove Themselves in Question Generation}

\author{Bang Nguyen \\
  University of Notre Dame \\
  \texttt{bnguyen5@nd.edu} \\\And
  Mengxia Yu \\
  University of Notre Dame \\
  \texttt{myu2@nd.edu} \\\AND 
  Yun Huang \\
  University of Illinois Urbana-Champaign \\
  \texttt{yunhuang@illinois.edu} \\\And 
  Meng Jiang \\
  University of Notre Dame \\
  \texttt{miiang2@nd.edu}
}

\begin{document}
\maketitle
\begin{abstract}
Reference-based metrics such as BLEU and BERTScore are widely used to evaluate question generation (QG).
In this study, on QG benchmarks such as SQuAD and HotpotQA, we find that using human-written references cannot guarantee the effectiveness of the reference-based metrics.
Most QG benchmarks have only one reference; we replicate the annotation process and collect another reference. A good metric is expected to grade a human-validated question no worse than generated questions. However, the results of reference-based metrics on our newly collected reference disproved the metrics themselves.
We propose a reference-free metric consisted of multi-dimensional criteria such as naturalness, answerability, and complexity, utilizing large language models.
These criteria are not constrained to the syntactic or semantic of a single reference question, and the metric does not require a diverse set of references.
Experiments reveal that our metric accurately distinguishes between high-quality questions and flawed ones, and achieves state-of-the-art alignment with human judgment.
\end{abstract}

\section{Introduction}

Question generation (QG) usually refers to the task of answer-aware question generation for controllability, aiming at generating a question based on a given context and answer span. Solutions are used to improve educational tools, build a product-based question-answering (QA) database, etc. Though anchored on a specific answer, there are still multiple ways of framing a question semantically and syntactically \cite{yu-jiang-2021-expanding, cho-etal-2019-mixture}. Users expect quality of every generated question.

\begin{figure*}[t]
    \centering
    \includegraphics[width=0.95\textwidth]{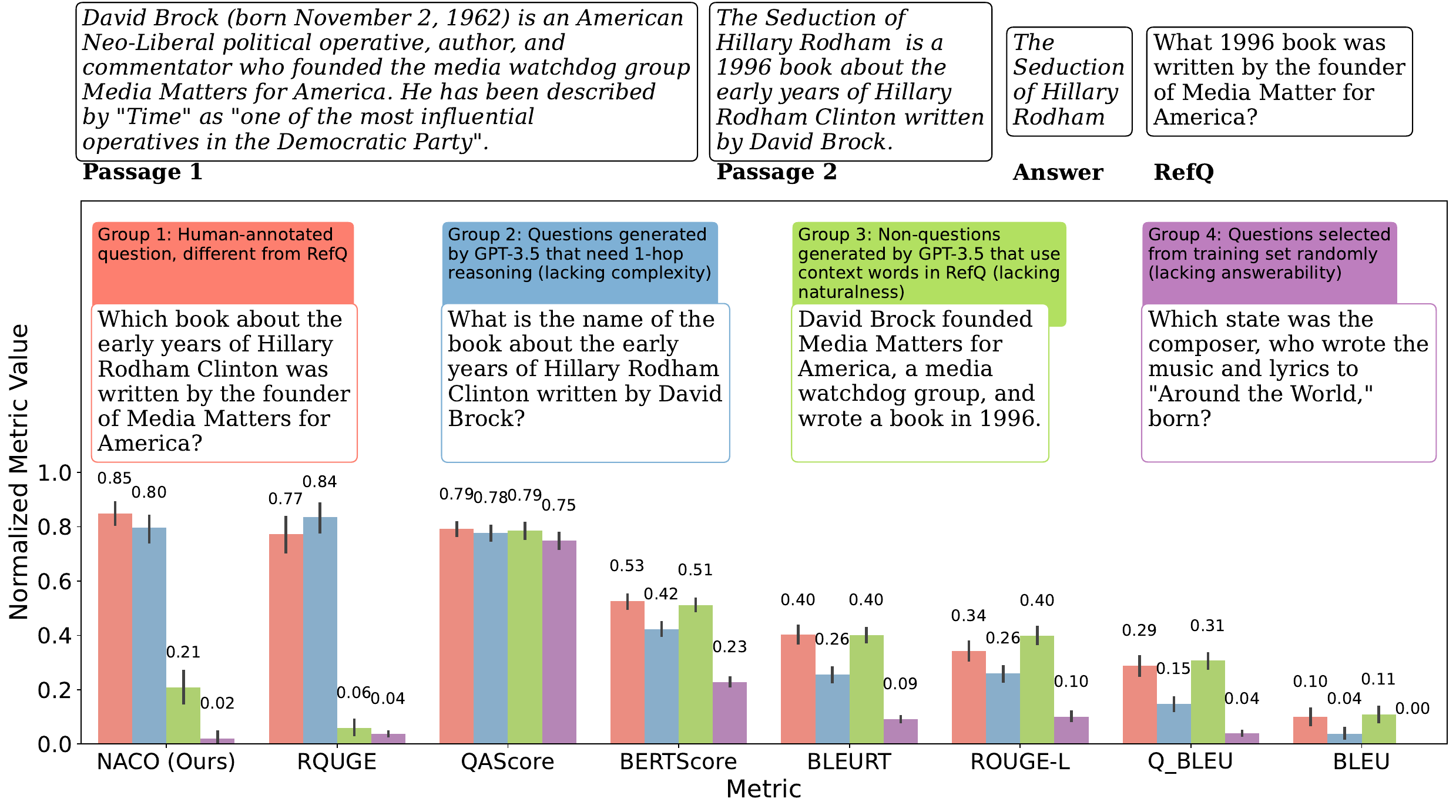}
    \caption{Normalized value of different evaluation metrics for four types of candidate questions against the same reference (RefQ) in the HotpotQA dataset \cite{yang2018hotpotqa}. Ideally, metrics should score Group 1 highest. Current QG metrics, except for NACo (ours) and RQUGE, primarily recognize random questions (Group 4) but fail to differentiate between Groups 1 and 3 (note the red and green bars). RQUGE, successfully identifies groups violating naturalness (Group 3) and answerability (Group 4), assigns a higher score for Group 2, which lacks complexity, than for Group 1. Our metric, shown in the leftmost bar group, prioritizing essential criteria of a question, can effectively distinguish all four groups of candidates while maintaining the highest rating for the valid questions.}
    \label{fig:metrics_comparison_graph}
\end{figure*}

To evaluate QG performance, reference-based metrics are widely used, which assess a machine-generated question against a human-written reference. The metrics are calculated either at the word level such as BLEU \cite{papineni2002bleu}, ROUGE \cite{lin2004rouge}, and METEOR \cite{banerjee-lavie-2005-meteor}, or in the embedding space such as BERTScore \cite{zhang2019bertscore}.
The challenges of using these evaluation metrics speak to the metrics themselves, considering word overlaps and/or semantic similarity between the generated question and the reference. In this sense, a QG model can ``cheat'' on the metrics by using many similar words to the reference, but ignoring essential components of a question. \citeauthor{mohammadshahi-etal-2023-rquge} questioned the effectiveness of reference-based metrics, developed a QA model, and defined a new metric named ``answerability'' or RQUGE. Though they showed a higher correlation with human preference, the failure of reference-based metrics was not studied, and the new metric's effectiveness is sensitive to the QA model's training and limited to its ability.

To disprove existing metrics, the challenge can be traced to the lack of diverse references for benchmark datasets. Previous works have shown that with access to a more diverse pool of references, the problem of poor correlation for these metrics can be mitigated \cite{freitag-etal-2020-bleu, oh-etal-2023-evaluation, tang2023metrics}. However, QG benchmarks often contain only one human-written ground-truth per example.

Our study starts from collecting another set of human-written references for two QG benchmarks,
following their standard annotation instructions. Besides the new references, we collect three groups of candidate questions, each lacking in an essential aspect of a question, for comparison.
We study how five reference-based metrics, namely {BLEU-4} \cite{papineni2002bleu}, {BLEURT} \cite{sellam-etal-2020-bleurt}, {ROUGE-L} \cite{lin2004rouge}, {BERTScore} \cite{zhang2019bertscore}, and {Q-BLEU} \cite{nema-khapra-2018-towards}, and two reference-free metrics, {QAScore} \cite{ji2022qascore}, and {RQUGE} \cite{mohammadshahi-etal-2023-rquge}, score the four groups of questions.
{Fig.~\ref{fig:metrics_comparison_graph} highlights the incompetency of current QG metrics in distinguishing the new reference (a valid question; see Group 1) from a less-complex-than-referenced question (Group 2), a non-question sentence that uses similar words (Group 3) or a randomly-selected question from training set (Group 4). Although these metrics tend to give higher scores for the new references than random questions, it remains challenging to separate them from the other less desirable candidates.}

Based on the above observations, we assert the failure of reference-based metrics in QG evaluation. We propose a shift to an evaluation mechanism that addresses essential criteria of a question that current metrics neglect: (1) \textbf{N}aturalness: \textit{how natural the question sounds}  \cite{wang-etal-2020-answer, bi-etal-2021-simple-complex}, (2) \textbf{A}nswerability: \textit{whether the question is grounded to the given answer}  \cite{ushio2022generative, ji2022qascore, nema-khapra-2018-towards, mohammadshahi-etal-2023-rquge}, and (3) \textbf{Co}mplexity: \textit{how likely it requires inferencing and synthesizing information} \cite{wang-etal-2020-answer, bi-etal-2021-simple-complex}. These criteria are not constrained to the syntactic and semantic structure of a single reference question. Thus, they address the challenges of evaluating question quality without access to a diverse set of references.

To overcome the limitation of the answerability measure in RQUGE \citep{mohammadshahi-etal-2023-rquge} and implement the other two measures, we utilize large language models (LLMs), which have demonstrated potential utility in data annotation tasks \cite{liu-etal-2023-g, wang2023aligning, lin-chen-2023-llm, chiang-lee-2023-large}, and their Chain-of-Thought (CoT) \cite{wei2022chain} process. We design CoT prompts for the LLM to directly measure the three criteria, as described in detail in $\S$\ref{sec:method}.


We name the three-dimensional metric \textbf{NACo}. The leftmost group of bars in Fig.~\ref{fig:metrics_comparison_graph} shows that NACo successfully distinguishes the valid questions (i.e., new human-written reference) from the other three groups with significant margins.
Reference-based metrics are so heavily influenced by the presence of overlapping words between the original reference and an invalid candidate that they even prefer the invalid candidate that NACo assigns a significantly lower score.

The key contributions of this paper include:
\begin{compactitem}
    \item We produce an additional set of human-written questions to current QG benchmarks, and show the unreliability of reference-based metrics in reflecting question quality.
    \item We propose NACo, a novel evaluation metric bridging the gap between human assessment and automated evaluation by assigning scores to three criteria of a good question.
    \item Through experiments and human evaluation, we demonstrate that NACo better aligns with human judgment of a good question than reference-based metrics for QG.
\end{compactitem}

We release the collected data and code implementation of NACo to facilitate future works. \footnote{\url{https://github.com/bnguyen5/naco}}



\section{Failure of Reference-based QG Metrics}
\label{sec:data_anno}


\subsection{Study Design \& Data Collection}

Previous studies questioning the effectiveness of reference-based metrics in QG typically rely on human evaluation. That is, they investigate whether the scores given to generated questions by QG metrics are highly correlated with the scores given by human evaluators \cite{mohammadshahi-etal-2023-rquge, ji2022qascore}. Unlike these studies, our research adopts a different approach during the data collection phase for QG datasets. Specifically, we replicate the data collection procedure of the datasets to collect new references, referred to as Group 1. Our focus is on determining if the newly collected references, when evaluated as candidates against the original references, receive high ratings from existing metrics. In addition, we extended our collection procedure to include three additional groups of candidate questions considered less desirable (Groups 2, 3, and 4) to ensure comprehensive comparisons. An effective and robust evaluation metric should assign a significantly higher score for questions in Group 1 compared to those in other groups.  Fig. \ref{fig:metrics_comparison_graph} illustrates our data collection process.



\textbf{Group 1: Human-written questions qualified as another reference for benchmark datasets}: We follow the procedure adopted by most papers collecting QA datasets. For each example to be annotated, we ask annotators, all fluent English speakers, to create a question based on some context passage(s) and a given answer \cite{rajpurkar2016squad}. If two passages are provided, we ask annotators to create a question such that it requires reasoning over both passages \cite{yang2018hotpotqa}.

\citeauthor{liu2020asking} proposed a concept of \textit{clues} for QG, which refers to words from the context passage that also appear in the question. Their experimental results indicate that the addition of a clue-prediction model enhances the performance of question generators on reference-based metrics. 
We investigate the usefulness of this concept by asking the annotators to phrase an additional question such that it contains the clue words used by the original annotators of the datasets. We ensure that the clue words are only presented to the annotators after they have finished creating their first question.

We perform the additional annotation on two popular QG benchmarks: (1) 748 test examples of SQuAD \cite{rajpurkar2016squad}, and (2) 96 test examples of HotpotQA \cite{yang2018hotpotqa}. To illustrate the application of our study, we collect another QG dataset in the educational domain, specifically from the TED-Ed learning platform\footnote{\url{https://ed.ted.com/}}. We further annotate 43 questions from this new dataset. More details about data collection and annotation for TedEdQA are provided in Appx. \ref{app:teded}.

For the HotpotQA sample, we also collect three other sets of questions, each violating an aspect required by the reference questions.

\textbf{Group 2: Single-hop questions for a multi-hop QG benchmark}: This group of candidate questions targets the multi-hop characteristic of HotpotQA where the ground-truth questions are formed based on two passages. Specifically, we select one from the original two passages that contains the answer span. We then ask GPT-3.5 to generate a question based on this single passage. We review the questions for grammar, clarity, relevance to the passage, independence from external knowledge, and a logical path to the answer. 


\textbf{Group 3: Non-questions that use the same words as the reference}: For this group of questions, we ask GPT-3.5 to generate a sentence based on the passages and use as many words from the same list of clues given to our annotators. We add a constraint such that the generated sentence cannot be in the form of a question. We then manually go through the generated sentences to ensure that no hallucinations were in place. 
In this sense, we produce a group of candidates that does not satisfy the most basic linguistic requirement of a question, naturalness, but still manages to contain many similar words as the ground-truth questions. 

\textbf{Group 4: Random questions from the training set}: The final set of candidate questions comes randomly from the training set of the benchmark. In the example illustrated in Fig. \ref{fig:metrics_comparison_graph}, the answer to this candidate question is \textit{Robin McLaurin Williams}, which is completely irrelevant to the given answer \textit{The Seduction of Hillary Rodham}. In this sense, this group of candidate questions violates the answerability aspect of an ideal candidate. 

\subsection{Results}



Fig.~\ref{fig:metrics_comparison_graph} shows the average normalized scores given by reference-based metrics to the four groups of candidate questions, all based on the same references. We find that all reference-based metrics, BLEU, ROUGE-L, BLEURT, Q-BLEU, and BERTScore, can effectively distinguish Group 4 (random questions) from the other groups, assigning it significantly lower scores. For instance, the average ROUGE-L score for Group 4 is $0.10$, compared to $0.34$ for Group 1, $0.26$ for Group 2, and $0.40$ for Group 3, with a minimum difference of $16\%$ from the scores of the other groups.

Fig.~\ref{fig:metrics_comparison_graph} also reveals issues with the reference-based metrics in accurately assessing Groups 1, 2, and 3. Notably, for all five reference-based metrics, Group 3, non-question sentences with wording similar to the references, receives the highest average score. For example, the ROUGE-L metric scores a non-question sentence that uses similar wording to the reference (green bar) on average $6\%$ higher than a new reference produced by our annotators (red bar), and $14\%$ higher than a perfectly answerable question requiring less reasoning than the reference (blue bar). This observation indicates a flaw in reference-based metrics, as candidates that do not form coherent questions should not receive higher scores than those that do.

The recently-introduced reference-free metrics, QAScore and RQUGE, also face difficulties in giving reasonable scores to questions from Groups 1, 2 and 3. QAScore, despite rating the new references highest among four groups, shows minimal score differences.
Meanwhile, RQUGE gives the highest average score ($0.84$) to Group 2, which contains single-hop questions in contexts requiring multi-hop reasoning. RQUGE's preference for single-hop questions can be attributed to its disregard for the complexity of the candidate question. It utilizes a pretrained QA model to compute a score based on the model's responses to the candidate question. The questions we collected, which require reasoning over two documents, may pose a greater challenge for the QA model compared to the simpler questions from Group 2. Since RQUGE's scoring mechanism does not consider the question's complexity, it underestimates the new references we collected in Group 1, scoring them at $0.77$.

Given the limitations of existing reference-based and reference-free metrics in accurately evaluating the four groups of questions, we propose a novel reference-free metric. This new metric aims to assess the quality of a question across multiple dimensions, providing a broader and more nuanced framework for assessing generated questions.

\section{NACo: A Novel Multi-dimensional Reference-free QG Metric}\label{sec:method}



Based on extensive review of the human evaluation procedure in QG literature, detailed in Appx. \ref{app:criteria}, we identify three essential criteria of a question: \textbf{N}aturalness, \textbf{A}nswerability, and \textbf{C}omplexity. We propose NACo, which leverages prompting and Chain-of-Thought (CoT) reasoning \cite{wei2022chain} to obtain a score for each criterion. Specifically, given the relevant context passage(s) and a question, we instruct an LLM as follows:
\begin{compactitem}
    \item The LLM first reads over the context passage(s) and the question. The LLM checks whether the question makes these mistakes: (1) not a question, (2) grammar errors, or (3) unclear objective. If so, the LLM should respond with `\textit{Question unnatural}', and we assign a score of $0$ for the question in terms of \textbf{naturalness} $n_{cand}$. Otherwise, $n_{cand}$ is $1$.
    \item Next, the LLM performs CoT reasoning to answer the question. Based on the LLM's CoT response, we obtain the \textbf{complexity} of the question by counting the number of reasoning steps the LLM made to answer the question.
    \item The LLM provides the final answer to the question. We define the \textbf{answerability} of the question $a_{cand}$ as the F1 score between the LLM's answer to the question and the ground-truth answer used to generate the question.
\end{compactitem}

The inherent qualities of questions speak to naturalness \cite{mohammadshahi-etal-2023-rquge} and answerability \cite{nema-khapra-2018-towards, ji2022qascore, mohammadshahi-etal-2023-rquge}, where higher values in these criteria indicate better quality in a question. We adopt a hierarchical scoring scheme that first examines the naturalness and answerability score obtained following the CoT-QA process. 
If the candidate question scores $0$ in these aspects, it is assigned a NACo score of $0$. 

If a candidate question passes the initial naturalness and answerability evaluation, we determine whether its complexity aligns with expected standards for the domain and dataset. For example, in the HotpotQA dataset, questions that require multi-hop reasoning might be preferred over simpler, single-hop questions. This preference may not hold in other datasets. In this sense, NACo relies on a subset of examples from the specific dataset to find the \textit{expected complexity} of a question in that dataset. Specifically, we perform the above CoT-QA process to obtain the complexity of the references. Expected complexity is then defined by the most common number of reasoning steps needed by the LLM to answer a reference question. In our experiments, we use 750 examples from the training set of SQuAD and HotpotQA to compute the expected complexity for each dataset.
Subsequently, NACo measures the similarity, denoted by $c_{cand}$, between the complexity of the candidate question and the expected complexity.

Overall, NACo is a weighted combination of $n_{cand}$, $a_{cand}$, and $c_{cand}$. In our experiments, we adopt a fair weight $\frac{1}{3}$ for each criterion. We provide additional details on how $c_{cand}$ is computed and integrated into the final score in Appx. \ref{app:naco_c}

\section{Experiments}
\begin{table}[!t]
    \begin{center}
    \resizebox{\linewidth}{!}{%
    \begin{tabular}{lrrrrr|r}
    \hline
    \noalign{\smallskip}
    \multirow{2}{*}{QG Competitor} & \multicolumn{5}{c|}{Ref-based metrics} & \multirow{2}{*}{NACo} \\
         & B & B-RT & R-L & BSc & Q-B  &
          \\ 
         \noalign{\smallskip}
         \hline
         \noalign{\smallskip}
        \textbf{LM-generated} &  &  &  &  &  &   \\ 
        BART-base & 19.53 & -0.28 & 44.79 & 92.13 & 36.94  &
        73.30\\ 
        GPT-3.5 (few-shot) & 18.06 & -0.23 & 43.58 & 92.18 & 36.48 & 
        73.67\\ 

        BART-clue-RefQ & \textbf{31.91} & \textbf{0.07} & \textbf{59.92} & \textbf{94.37} & \textbf{52.33} & 
        69.97 \\ 
        \noalign{\smallskip}
        \hline
        \noalign{\smallskip}
        \textbf{Human-validated} &  &  &  &  &  &    \\ 
        RefQ & 100.00 & 1.00 & 100.00 & 100.00 & 100.00 &  
        \textbf{75.09} \\ 
        AnnoQ & 12.78 & -0.31 & 37.83 & 91.52 & 31.32 & 
        74.01 \\ 
        AnnoQ-clue-RefQ & \underline{27.43} & \underline{0.04} & \underline{53.62} & \underline{93.85} & \underline{46.89} &
        \underline{74.21}\\ 
        \noalign{\smallskip} \hline
    \end{tabular}}
    \caption{\label{result_squad_refq}\textbf{SQuAD} - Performance of different QG methods on NACo and other existing metrics. The evaluation uses original SQuAD questions (RefQ) as references, with GPT-3.5 as the underlying LLM. The highest and second-highest scores (not including references for reference-based metrics) are highlighted with bold and underline markers, respectively.}
    \end{center}
\end{table}

 \begin{table*}[!t]
    \centering
    \resizebox{0.95\textwidth}{!}{%
     \begin{tabular}{l|rrr|rrr|rrr|rrr}
    \hline
    \noalign{\smallskip}
     \multirow{2}{*}{Metric} & \multicolumn{3}{c|}{Naturalness} & \multicolumn{3}{c|}{Answerability} & \multicolumn{3}{c|}{Complexity} & \multicolumn{3}{c}{\textbf{Overall}} \\ 
     & $ r $ & $ \rho $ & $ \tau $ & $ r $ & $ \rho $ & $ \tau $ & $ r $ & $ \rho $ & $ \tau $ & $ r $ & $ \rho $ & $ \tau $ \\ \noalign{\smallskip} \hline \noalign{\smallskip}
    \textbf{Ref-based metric} &  &  &  &  &  &  &  &  &  &  & & \\ 
     \hspace{0.5cm}B &   0.08 &    -0.06 &    -0.04 &   0.16 &    0.06 &   0.05 &   0.40 &     0.38 &       0.31 &   0.25 &     0.20 &   0.16 \\
     \hspace{0.8cm}w/ \textit{DivRef} &   0.09 &    -0.01 &    -0.01 &   0.18 &    0.10 &   0.08 &   0.43 &     0.42 &       0.33 &   0.27 &     0.25 &   0.19 \\
     \hspace{0.5cm}B-RT &     0.12 &     0.03 &    0.03 &    0.23 &     0.19 &    0.14 &    0.49 &     0.50 &    0.37 &    0.33 &     0.35 &    0.24 \\
     \hspace{0.8cm}w/ \textit{DivRef} &     0.14 &     0.05 &    0.04 &    0.26 &     0.22 &    0.16 &    0.51 &     0.52 &    0.39 &    0.36 &     0.38 &    0.26 \\
     \hspace{0.5cm}R-L &    0.07 &     -0.03 &    -0.03 &    0.18 &     0.10 &    0.07 &    0.43 &     0.42 &     0.31 &    0.27 &     0.24 &    0.17 \\
      \hspace{0.8cm}w/ \textit{DivRef} &    0.13 &     0.06 &    0.04 &    0.24 &     0.19 &    0.15 &    0.48 &     0.48 &     0.37 &    0.34 &     0.34 &    0.24 \\
     \hspace{0.5cm}BSc &    0.18 &     0.09 &    0.06 &     0.27 &     0.21 &    0.16 &    0.52 &      \underline{0.54} &    \underline{0.41} &    0.38 &     0.39 &    0.27 \\
     \hspace{0.8cm}w/ \textit{DivRef} &    0.23 &     0.14 &    0.10 &     0.33 &     0.29 &    0.21 &    \textbf{0.56} &      \textbf{0.57} &    \textbf{0.43} &    0.44 &     0.45 &    0.31 \\
    \hspace{0.5cm}Q-B &    0.07 &     -0.05 &    -0.04 &    0.17 &     0.09 &    0.07 &    0.45 &     0.44 &    0.32 &    0.27 &     0.26 &    0.17 \\
    \hspace{0.8cm}w/ \textit{DivRef} &    0.07 &     -0.04 &    -0.03 &    0.16 &     0.09 &    0.07 &    0.45 &     0.44 &    0.32 &    0.27 &     0.26 &    0.18 \\
    \noalign{\smallskip} \hline \noalign{\smallskip}
    \textbf{Ref-free metric} &  &  &  &  &  &  &  &  &  &  & & \\ 
    \hspace{0.5cm}QA-S &    -0.01 &     0.00 &    0.00 &    0.01 &     -0.01 &    -0.01 &    0.04 &     0.03 &    0.02 &    0.02 &    0.02 &    0.02 \\
    \hspace{0.5cm}R-Q &    0.38 &     0.26 &    0.20 &    0.67 &   0.51 &    0.39 &    0.33 &     0.22 &     0.16 &    0.57 &     0.38 &    0.27 
    \\
    \noalign{\smallskip} \hline \noalign{\smallskip}
    \textbf{NACo (Ours)} &  &  &  &  &  &  &  &  &  &  & & \\ 
    \hspace{0.5cm}Llama3-8B &     \cellcolor{customgreen!30}0.49 &     \cellcolor{customgreen!30}\underline{0.32} &    \cellcolor{customgreen!30}\underline{0.25} &    0.66 &     0.50 &    0.39 &    0.43 &     0.36 &    0.21 &    \cellcolor{customgreen!30}0.64 &      \cellcolor{customgreen!30}0.42 &    0.30 \\
    \hspace{0.5cm}Mixtral-8x7B &     \cellcolor{customgreen!30}0.36 &     \cellcolor{customgreen!30}0.30 &    \cellcolor{customgreen!30}0.23 &    0.54 &     \cellcolor{customgreen!30}0.52 &    \cellcolor{customgreen!30}0.40 &    0.36 &     0.23 &    0.17 &    0.56 &      \cellcolor{customgreen!30}0.40 &    0.28 \\
    \hspace{0.5cm}Claude-Haiku &     \cellcolor{customgreen!30}\underline{0.53} &     \cellcolor{customgreen!30}0.30 &    \cellcolor{customgreen!30}0.23 &    \cellcolor{customgreen!30}\underline{0.71} &     0.51 &    \cellcolor{customgreen!30}0.40 &    0.50 &     0.35 &    0.27 &    \cellcolor{customgreen!30}\underline{0.71} &      \cellcolor{customgreen!30}0.47 &    \cellcolor{customgreen!30}\underline{0.35} \\
    \hspace{0.5cm}GPT3.5 &     \cellcolor{customgreen!30}0.47 &     \cellcolor{customgreen!30}0.30 &    \cellcolor{customgreen!30}0.23 &    \cellcolor{customgreen!30}0.70 &     \cellcolor{customgreen!30}\underline{0.53} &    \cellcolor{customgreen!30}\underline{0.41} &    0.49 &     0.35 &    0.27 &    \cellcolor{customgreen!30}0.68 &     \cellcolor{customgreen!30}\underline{0.48} &    \cellcolor{customgreen!30}\underline{0.35} \\
    \hspace{0.5cm}GPT4 &     \cellcolor{customgreen!30}\textbf{0.64} &     \cellcolor{customgreen!30}\textbf{0.36} &    \cellcolor{customgreen!30}\textbf{0.28} &    \cellcolor{customgreen!30}\textbf{0.78} &     \cellcolor{customgreen!30}\textbf{0.59} &    \cellcolor{customgreen!30}\textbf{0.47} &    \underline{0.55} &     0.35 &    0.27 &    \cellcolor{customgreen!30}\textbf{0.80} &      \cellcolor{customgreen!30}\textbf{0.52} &    \cellcolor{customgreen!30}\textbf{0.39} \\
      \noalign{\smallskip} \hline
    \end{tabular}}
    \caption{\label{table:human-correlation} Correlation between human assessments and automated evaluation metrics as indicated by Pearson $r$, Spearman $\rho$, and Kendall $\tau$ correlation coefficients. For reference-based metrics, we report the metric's correlation with human judgment both when only the original reference is used and when adding the diversified references (w/ \textit{DivRef}). The highest and second-highest scores are highlighted with bold and underline markers, respectively. Shaded regions indicate an improvement compared to the current state-of-the-art metric for that respective column.}
\end{table*}

\subsection{Experimental Setup}

\textbf{Question generation competitors}: We compare the evaluation capacity of NACo with that of current QG metrics on four QG models and three sets of human-validated references. \textit{Generative Language Models} like BART \cite{lewis2020bart} and T5 \cite{raffel2020exploring} are current state-of-the-art QG performers on reference-based metrics \cite{ushio2022generative}. We fine-tune BART-base using the training set, following the method introduced by \citeauthor{chan-fan-2019-recurrent}. We also produce another version of BART-base, \textbf{BART-clue-RefQ}, which highlight the ground-truth clues used by reference questions (RefQ) in the context given as input to BART-base (detailed in \ref{app:exp_details}). In addition, we use GPT-3.5 to generate questions for the test examples through zero-shot, and few-shot prompting. In the few-shot setting, we randomly select 10 examples from the training set of the dataset as demonstrations.
Alongside the original reference questions provided by the datasets (\textbf{RefQ}), we use the annotated data detailed in $\S$\ref{sec:data_anno} to obtain two human-validated competitors: \textbf{AnnoQ}, which contains the questions written by our annotators before given gold clues, and \textbf{AnnoQ-clue-RefQ}, which contains the gold-clue-guided questions written by our annotators.

\textbf{Baselines}: We compare the evaluation capacity of NACo with five reference-based metrics, including {BLEU-4} (B) \cite{papineni2002bleu}, {BLEURT} (B-RT) \cite{sellam-etal-2020-bleurt}, {ROUGE-L} (R-L) \cite{lin2004rouge}, {BERTScore} (BSc) \cite{zhang2019bertscore}, and {Q-BLEU} (Q-B) \cite{nema-khapra-2018-towards}, and two reference-free metrics, {QAScore} (QA-S) \cite{ji2022qascore}, and {RQUGE} (R-Q)\cite{mohammadshahi-etal-2023-rquge}. \citeauthor{tang2023metrics} proposes using LLM to diversify the limited references in benchmarks, demonstrating an improvement in the correlation between reference-based metrics and human judgment. We replicate this approach and report the evaluation performance of the five reference-based metrics both when only the original reference is used and when adding the diversified references.

\textbf{NACo implementation}: We provide the CoT prompt used in our experiments in Appx. \ref{app:prompt}. We experimented with five underlying LLMs: Llama3-8B, Mixtral-8x7B, Claude3-Haiku, GPT3.5-turbo, and GPT4o.

\textbf{Human Evaluation}: We recruit volunteer annotators, all fluent English speakers, to evaluate both model-generated questions and human-written questions, using 96 test examples from HotpotQA. For each example, annotators evaluate four questions: RefQ, GPT-3.5 (zero-shot), BART-base, and AnnoQ, displayed in randomized and anonymized order. Evaluators rate each question based on naturalness, answerability, and complexity, using a 3-point scale for each criterion. Additionally, we sum the individual scores to calculate a combined score that reflects the question's overall quality. We obtain three annotations per question and use the average of these as the standard for human judgment. The Pearson correlation coefficient between ratings given by our annotators is 0.67. The rating rubric is available in Appx. \ref{app:survey_instructions}.

\subsection{Results}

\begin{figure*}[t]
  \begin{minipage}[b]{0.48\textwidth}
    \centering
    \includegraphics[width=\linewidth]{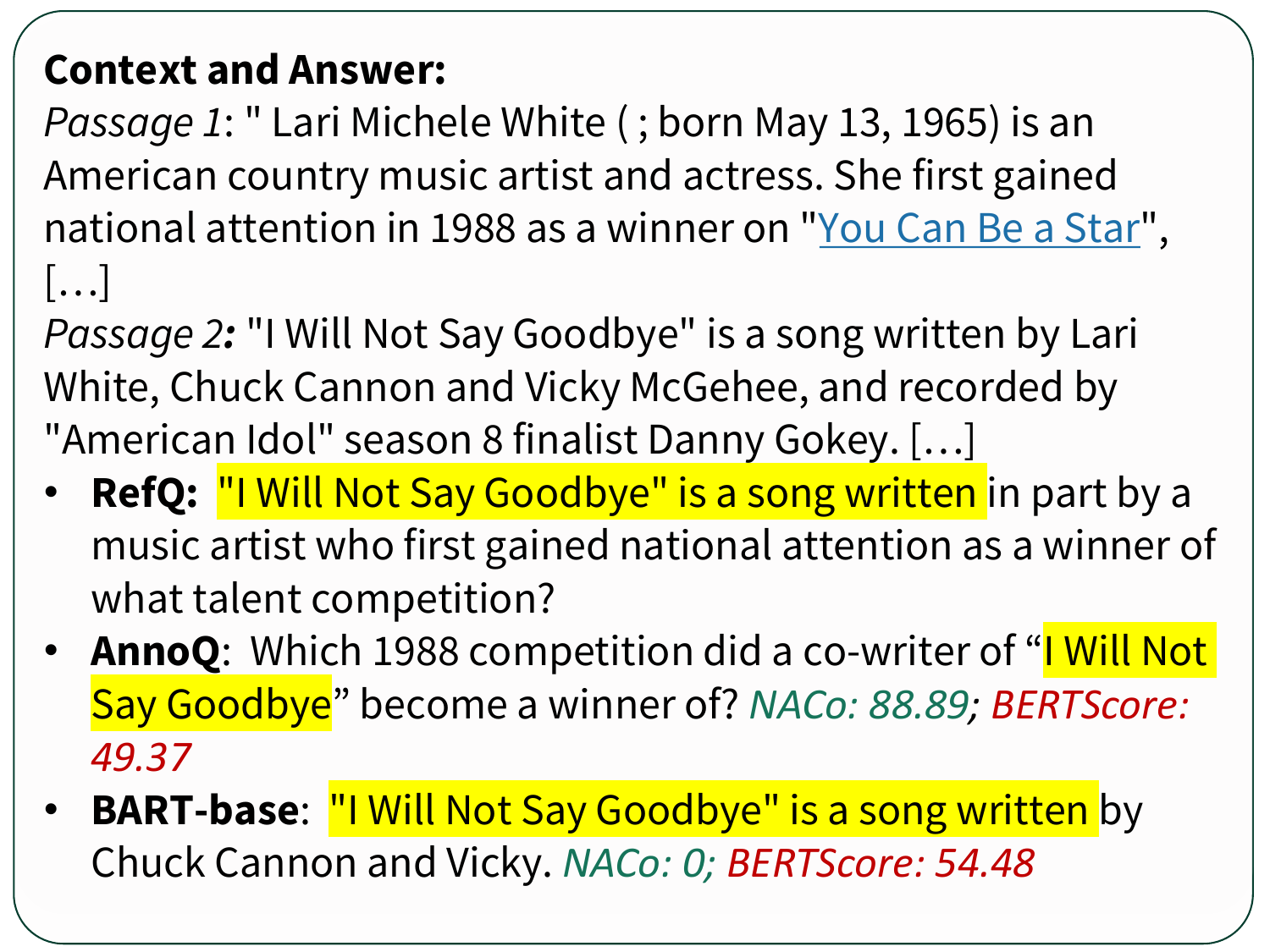}
    \caption{Case study 1: NACo vs BERTScore. Longest common subsequences between candidate question and RefQ are highlighted.}
    \label{fig:naco_vs_bertscore}
  \end{minipage}
  \hfill
  \begin{minipage}[b]{0.48\textwidth}
    \centering
    \includegraphics[width=\linewidth]{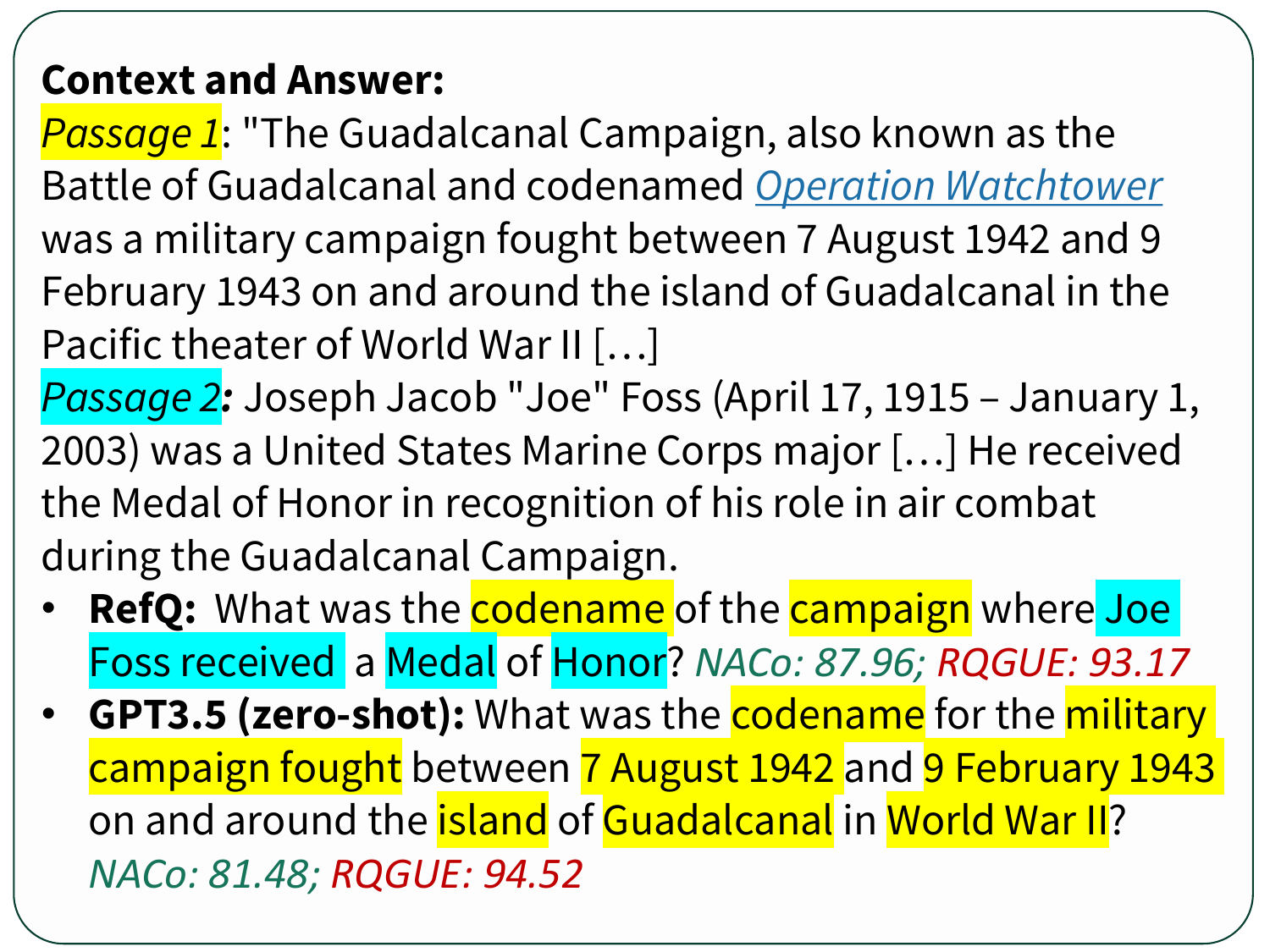}
    \caption{Case study 2: NACo vs RQUGE. Context words used by the question are highlighted in the same color if they come from the same passage.}
    \label{fig:naco_vs_rquge}
  \end{minipage}
\end{figure*}

\textbf{Failure of reference-based metrics}:
We report QG competitors' performance on various metrics, including NACo, using RefQ as the reference in Tbl. \ref{result_squad_refq} for SQuAD.
Even though RefQ, AnnoQ, and AnnoQ-clue-RefQ are all qualified as valid questions, reference-based metrics rate them with significant differences. In the SQuAD dataset, BLEU scores for RefQ, AnnoQ, and AnnoQ-clue-RefQ are 100, 12.78, and 27.43, respectively (Tbl. \ref{result_squad_refq}). However, NACo rates these three groups of questions similarly, with RefQ, AnnoQ, and AnnoQ-clue-RefQ scoring 75.09, 74.01, and 74.21, respectively (Tbl. \ref{result_squad_refq}). Similar patterns are observed in the HotpotQA and TedEdQA datasets, as detailed in Tbl. \ref{result_hotpot_refq} and Tbl. \ref{result_teded_refq}.

According to reference-based metrics, models that learn from training data either through finetuning (like BART-base) or demonstration (like GPT-3.5) are scored significantly higher than our annotators, who lack access to the training data. For instance, in the case of SQuAD, BART-base is scored higher than AnnoQ by almost $7\%$ according to BLEU-4, reported in Tbl. \ref{result_squad_refq}. As reference-based metrics measure syntactic and semantic similarity, the use of a single reference can disqualify our annotated questions from being considered reference materials, resulting in a misleading portrayal of a valid group of candidate questions.

\textbf{Effectiveness of NACo}: Referring to our analysis of four groups of candidate questions for HotpotQA in Fig. \ref{fig:metrics_comparison_graph}, NACo uniquely succeeds in separating all four groups by significant margins, unlike the seven existing metrics. The newly collected multi-hop questions in Group 1, which satisfy all criteria for HotpotQA questions, achieve the highest average NACo score of $0.85$. They are followed by the questions in Group 2, lacking in complexity, with a score of $0.80$; Group 3, lacking in naturalness, with a score of $0.21$; and Group 4, lacking in answerability, with a score of $0.02$.

We calculate the Pearson $r$, Spearman $\rho$, and Kendall $\tau$ correlation coefficients to measure the agreement between all metrics, including NACo, and human judgment, as reported in Tbl. \ref{table:human-correlation}. This comparison considers correlation with both individual criteria and the overall question quality. Tbl. \ref{table:human-correlation} reveals that NACo demonstrates the highest correlation with human evaluation for individual criteria in 9 out of 12 scores. Notably, NACo exhibits the strongest agreement with human judgment concerning the overall quality of questions across all correlation metrics. This observation is consistent across different underlying LLMs.

\subsection{Analysis}
\begin{table}[h]
    \begin{center}
    \resizebox{\linewidth}{!}{%
     \begin{tabular}{lr|r|r|r}
    \hline
    \noalign{\smallskip}
        QG Competitor & B & R-Q & NACo & Human\\ \noalign{\smallskip} \hline \noalign{\smallskip}
        \textbf{LM-generated} &  &  &  \\ 
        BART-base & 14.57 & 2.90 & 42.65 & 2.80\\ 
        GPT-3.5 (zero-shot) & 9.46 & 4.18 & 74.99 & 4.60\\ 
        \noalign{\smallskip} \hline \noalign{\smallskip}
        \textbf{Human-validated} &  &  &  \\ 
        RefQ & 100.00 & 4.12 & 75.32& 5.14\\ 
        AnnoQ & 14.80 & 4.21 & 84.97 & 5.45 \\ 
        \noalign{\smallskip} \hline
    \end{tabular}}
    \caption{\label{result_hotpot_refq_small} \textbf{HotpotQA} -  Performance of different QG methods on NACo and other existing metrics. The evaluation uses original HotpotQA questions (RefQ) as references, with GPT-3.5 as the underlying QA system for NACo.}
    \end{center}
\end{table}

\textbf{NACo vs. Reference-based Metrics}: Tbl. \ref{result_hotpot_refq_small} indicates that reference-based metrics rate BART-base questions slightly lower than AnnoQ (by $0.23\%$ according to BLEU), whereas NACo shows a much larger gap ($42.32\%$). Upon manually reviewing the questions generated by BART-base, we noticed a considerable number of them were not actual questions but rather statements using similar wording to the reference question RefQ. This observation is validated by our human evaluators, detailed in \ref{app:human_eval_results}. Fig. \ref{fig:naco_vs_bertscore} provides a case study where BERTScore, the reference-based metric most aligned with human judgment \cite{mohammadshahi-etal-2023-rquge}, favored BART-base generation over the human annotated question, even though the former was not formatted as a question. This incompetency of reference-based metric can be explained by the fact that BART, when finetuned on the HotpotQA training set, can identify words that will be used in the reference RefQ,  but fail to form a coherent and answerable question. NACo, emphasizing essential criteria of a question, assigns a score of $0$ to the BART-base output while giving a high score for AnnoQ ($88.89$).

\begin{figure*}[t]
    \centering
    \includegraphics[width=0.95\textwidth]{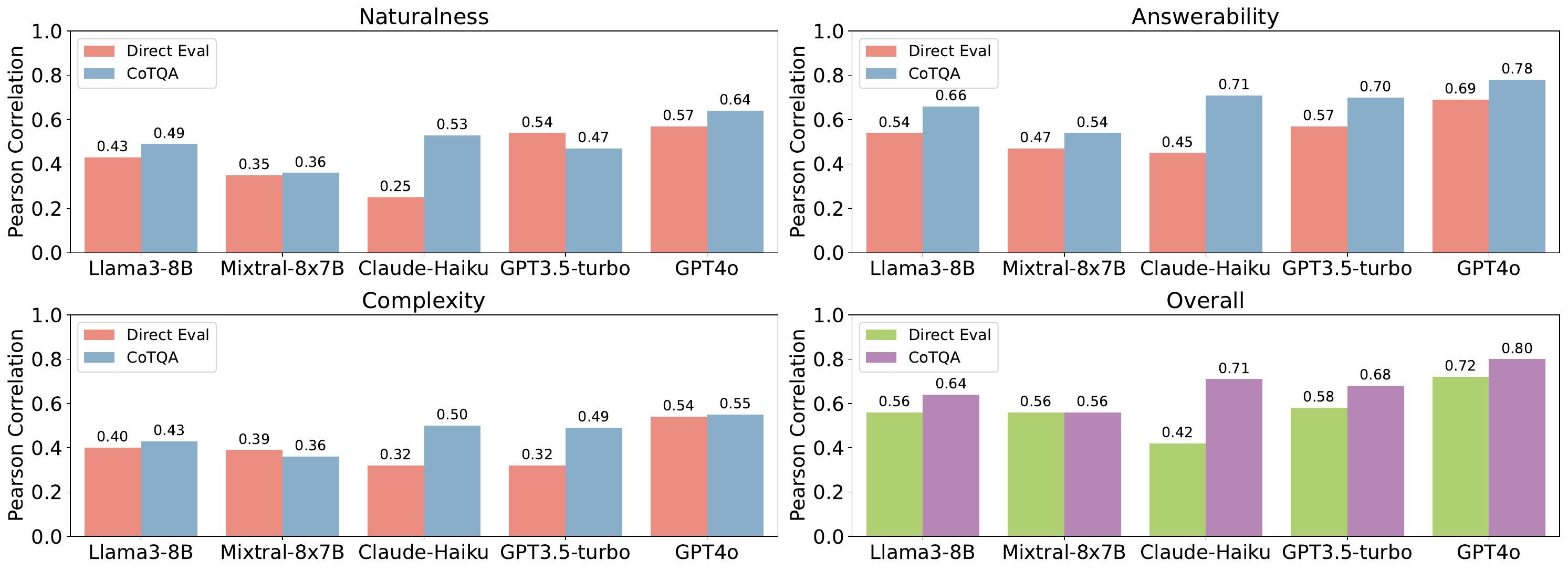}
    \caption{Correlation with human judgement - Comparing CoT-QA (NACo) with Direct Evaluation}
    \label{fig:direct_vs_cot}
\end{figure*}

\textbf{NACo vs. Existing Reference-free Metrics}: Tbl. \ref{result_hotpot_refq_small} also reveals that the new reference-free metric for QG, RQGUE, rates GPT-3.5 generated questions—whether in zero-shot or few-shot modes—comparably to the original reference question (RefQ). A manual review showed that GPT-3.5 typically utilizes only one of two context passages for creating a multi-hop question, as illustrated in Fig. \ref{fig:naco_vs_rquge}. Again, human evaluation verifies our observations, detailed in Appx. \ref{app:human_eval_results}. In the case study, GPT-3.5 exclusively used context words from Passage 1, making access to Passage 2 unnecessary for answering the question. Meanwhile, RefQ incorporates context words from both passages and requires reasoning across both for an answer. RQUGE overlooks this aspect and assigns a higher score for the GPT-3.5 question than for RefQ ($94.52$ and $93.17$, respectively). 
Addressing this gap, NACo acknowledges the answerability and naturalness of the GPT-3.5 question, but penalizes its lower-than-expected complexity, resulting in a score of $81.48$. Since RefQ meets all three criteria of a candidate question, NACo awards it a higher score of $87.96$.

\textbf{Human Preference Study}: As our human evaluation study assesses candidate questions based on the criteria that NACo measures, we conduct a human preference study to ensure fair comparisons between NACo and existing metrics. In this study, we further compare NACo with the top two baselines: RQUGE and BERTScore. We sample 20 pairs of questions where NACo and each baseline disagree in their scoring. Three human evaluators make their preferences for each pair, achieving a Cohen's Kappa of 0.87, indicating strong agreement. Using human preference as the reference, NACo wins against RQUGE in 15 out of 20 cases (75\%) and BERTScore in 12 out of 20 cases (60\%).

\textbf{NACo (CoT-QA) vs. LLM Direct Evaluation}: Large language models (LLMs) are increasingly utilized as proxies for human evaluators. Previous studies have suggested that when receiving CoT instructions typically given to human evaluators, LLMs can assess generated texts in a way that are highly aligned with human judgement \cite{liu-etal-2023-g}. We refer to this approach of using LLM evaluators as Direct Evaluation (DirectEval). We examine the effectiveness of CoT-QA, used by NACo, against DirectEval, which provides the LLMs the same human evaluation instructions in Appx. \ref{app:survey_instructions}. Fig. \ref{fig:direct_vs_cot} presents the Pearson correlation coefficients, comparing the performance of CoT-QA (NACo) with DirectEval across individual criteria and overall question quality. The results indicate a higher alignment with human judgment when employing CoT-QA for each respective LLM.  Notably, adopting CoT-QA instead of DirectEval significantly boosts the performance of Claude-Haiku, improving the alignment with human judgment of overall question quality from $0.42$ to $0.71$. This improvement is comparable to the performance achieved using GPT4o ($0.72$ in DirectEval setting, $0.80$ in CoTQA setting), while being $12$ times more cost effective.

We also investigate whether DirectEval and NACo will benefit from the addition of a reference question during evaluation. For DirectEval, we provide the reference question at the end of the instruction. For NACo, we use the reference question's complexity (obtained from CoT-QA) as the expected complexity. In our experiment, we use the RefQ group as the reference, and evaluate the other three groups of candidate questions (BART-base, GPT3.5, and AnnoQ). Table \ref{analysis_ref_naco} details the correlation between human judgement and the use of LLM evaluators with and without references.

\begin{table}[h]
    \begin{center}
    \small
     \begin{tabular}{lr|r}
    \hline
    \noalign{\smallskip}
        Method & GPT3.5 & Claude-Haiku\\ \noalign{\smallskip} \hline \noalign{\smallskip}
        DirectEval & 0.63 & 0.48 \\ 
        DirectEval + RefQ & 0.74 & 0.52 \\ 
        \noalign{\smallskip} \hline \noalign{\smallskip}
        NACo & 0.72 & 0.74\\ 
        NACo + RefQ & 0.69 & 0.75 \\ 
        \noalign{\smallskip} \hline
    \end{tabular}
    \caption{\label{analysis_ref_naco} Pearson correlation between human assessments and LLM evaluators, with and without references (RefQ).
    }
    \end{center}
\end{table}

It can be seen that while references benefit DirectEval to some extent, they do not consistently improve it to the level of NACo. Specifically, DirectEval $+$ RefQ with GPT3.5 shows on-par performance with NACo (0.74 v.s. 0.72), while DirectEval $+$ RefQ with Claude-Haiku underperforms NACo by a large margin. Furthermore, adding references on top of NACo does not further improve its performance. These results suggest that a single reference does not provide orthogonal benefits to NACo, which aligns with our findings regarding the limitations of reference-based metrics. NACo, as a reference-free metric, provides comprehensive and robust QG evaluations, without suffering from the bias of a single reference and the expensive reference collection process.

\section{Related Work}

\textbf{Evaluation Metrics for Question Generation}:
The evaluation of QG models commonly used reference-based metrics such as BLEU \cite{papineni2002bleu}, ROUGE \cite{lin2004rouge}, METEOR \cite{banerjee-lavie-2005-meteor}, BERTScore \cite{zhang2019bertscore}, and BLEURT \cite{sellam-etal-2020-bleurt}. Based on correlation with human judgment, there have been studies attempting to challenge the effectiveness of these reference-based metrics and propose reference-free evaluation mechanism for QG \cite{nema-khapra-2018-towards, ji2022qascore, mohammadshahi-etal-2023-rquge}. Our study, on the other hand, questions the competency of reference-based metrics by replicating the data collection process of benchmarks and introducing new references. Other works have taken a similar approach, designing and collecting different groups of candidates to investigate reference-based metrics in machine translations \cite{amrhein-etal-2022-aces, karpinska-etal-2022-demetr} and question answering \cite{bulian-etal-2022-tomayto}. However, QG poses unique challenges to the evaluation of question quality, considering aspects such as complexity and answerability, and therefore call for a study like ours.

\textbf{LLMs as evaluators for NLG tasks}: A growing research interest revolves around the use of large language models (LLMs) for evaluating quality of generated texts \cite{liu-etal-2023-g, wang2023aligning, lin-chen-2023-llm, chiang-lee-2023-large}. Investigating GPT-3 and its variances' evaluation capacity on story generation and adversarial attack tasks, \citeauthor{chiang-lee-2023-large} found that when given the same instructions as human annotators, LLMs show positive correlation with human judgment. \citeauthor{lin-chen-2023-llm} and \citeauthor{liu-etal-2023-g} obtained similar observations for dialogue generation and text summarization tasks. Due to the recent nature of this research direction, no other work has performed a comprehensive study on the use of LLMs as evaluators for the question generation task.

\section{Conclusion}
In this work, we questioned the competency of reference-based metrics in providing an accurate assessment for question generation. We replicated the data collection process used for benchmark datasets, gathering candidate questions qualified as new references. Our analysis highlights the shortcomings of reference-based metrics in differentiating new references from flawed candidates, assigning significantly lower scores to the former. Even the recently introduced reference-free metric, RQUGE, face difficulties in this regard. To address these challenges, we introduce NACo, a multi-dimensional, reference-free metric bridging the gap between automated evaluation and human judgment in question generation. Our experimental results showcase that NACo, leveraging the Chain-of-Thought capabilities of Large Language Models for question answering, not only meets the expectations for quantitative QG metrics but also achieves state-of-the-art alignment with human evaluation.

\section*{Limitations}

A limitation of our work speaks to the required access to a reasonable number of references to assess domain-specific or dataset-specific complexity. Future works can investigate how to account for expected complexity in scenarios where references are limited and difficult to collect. Moreover, NACo, like other reference-free metrics for QG, is subject to the performance of the underlying QA model. Specifically, the constraints of GPT-3.5 in answering complex, multi-hop questions might have limited NACo's ability to evaluate valid references closer to the upperbound. We provide a case study to illustrate this issue in Appx \ref{app:error_analysis}. Future directions should explore evaluation frameworks that are robust to variations in QA model performance.

\section*{Acknowledgements}
This work was supported by NSF IIS-2119531, IIS-2137396, IIS-2142827, IIS-2234058, CCF-1901059, and ONR N00014-22-1-2507.


\bibliography{custom}

\appendix

\section{Appendix}
\label{sec:appendix}


\subsection{What makes a good question?}
\label{app:criteria}

After the best model for question generation has been developed, it often goes through a round of human evaluation to assess the quality the generated questions. The human evaluation stage often looks at the following aspects of the generated question:

\textbf{Naturalness} \cite{wang-etal-2020-answer, bi-etal-2021-simple-complex} addresses essential linguistic elements of a question, such as whether the question is free from grammar mistakes \cite{ushio2022generative}, or how clear and fluent the question sounds \cite{pan-etal-2020-semantic, laban-etal-2022-quiz}.

\textbf{Answerability} measures how well the question is grounded to the input context and answer. In this sense, a good question should be relevant to the input context \cite{pan-etal-2020-semantic, wang-etal-2020-answer}, and in the answer-aware setting, should have a reasoning path that leads to the given answer \cite{ushio2022generative, ji2022qascore, nema-khapra-2018-towards, mohammadshahi-etal-2023-rquge}. 

\textbf{Complexity} \cite{wang-etal-2020-answer, bi-etal-2021-simple-complex} speaks to the reasoning path taken to answer the question. The higher number of reasoning steps needed, the more complex the question. It should be noted that higher complexity does not necessarily indicate better quality in a question. This quality rather depends on the nature of the dataset.

\begin{figure}[t]
    \centering
    \includegraphics[width=\linewidth]{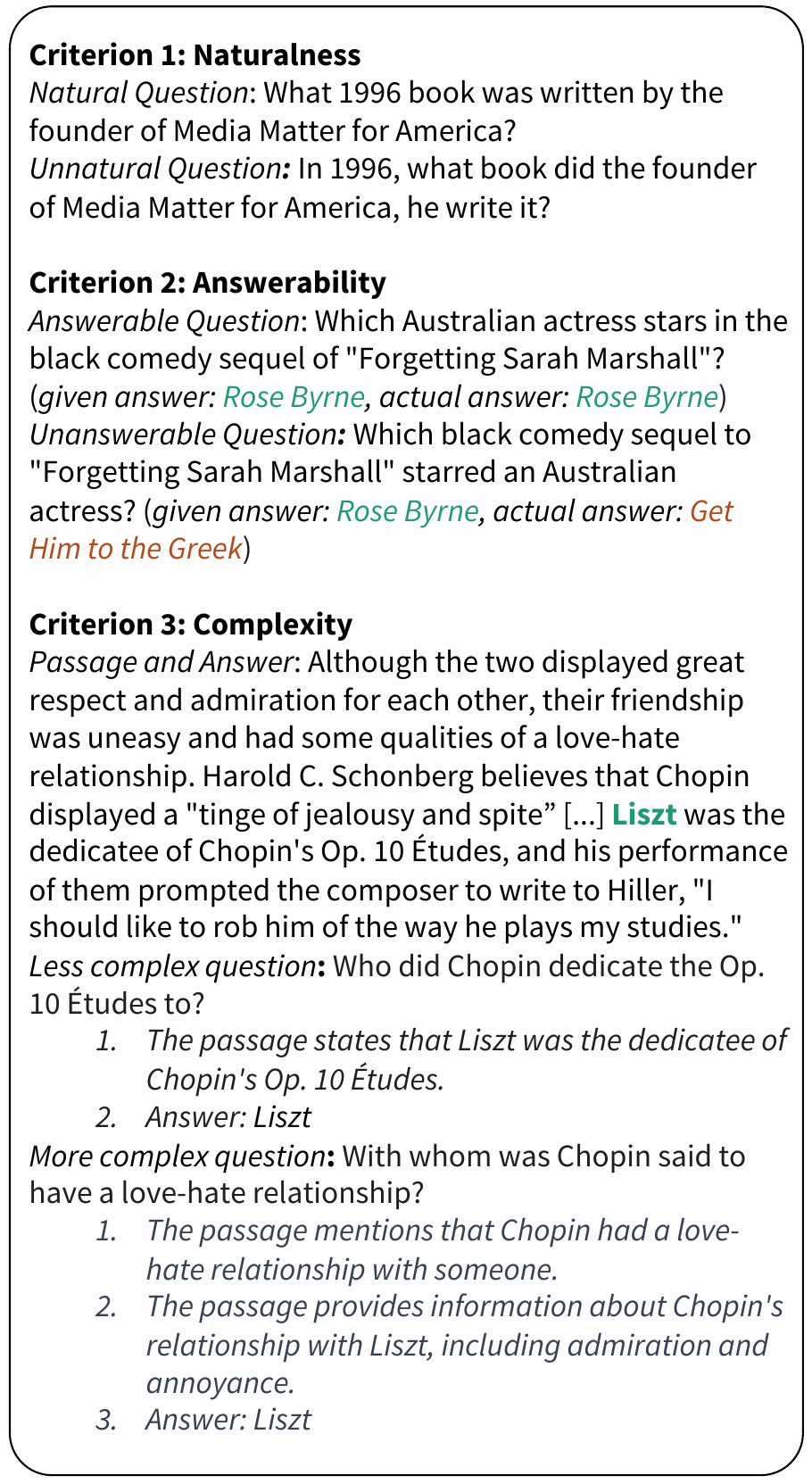}
    \caption{Examples for each criterion addressed by our metric: Naturalness, Answerability, and Complexity.}
    \label{fig:naco_examples}
\end{figure}

Fig. \ref{fig:naco_examples} illustrates the gap between current automatic QG metrics and human evaluated metrics, where two questions using similar words can have opposite qualities. This gap can be explained by the fact that existing automatic metrics do not directly address any of the criteria that human annotation often looks for in a question. To address this challenge, our metric integrates the human perspective of a "good" question: naturalness, answerability, and complexity, into the evaluation pipeline.


\subsection{Prompt for CoT-QA}\label{app:prompt}
You will be given [one/two] context passage(s) and a sentence. If the sentece is a question, your task is to output a text span from the context passage to answer the question. Your answer should NOT be complete sentences.

\noindent Instructions:
\begin{enumerate}
    \item Let's read the passage first and then read the sentence. Consider:
    \begin{enumerate}
        \item Is the sentence a question? If yes, what information indicates that it is a question? If not, output `not a question' and stop generation.
        \item If it is a question, considers if the question is unclear, or has grammar errors. If so, output `Question unnatural'.
    \end{enumerate}
    \item Now find the answer to the question. Speak out loud your detailed reasoning.
    \item Highlight your answer between two <ans> tokens.
\end{enumerate}

\noindent Format you response as follows:
\begin{enumerate}
    \item Your response to 1a and 1b
    \item Step by step reasoning:
    \begin{enumerate}
        \item Step 1 [reasoning step must be a single sentence with one clause]
        \item Step 2 [reasoning step must be a single sentence with one clause]
        \item ...
    \end{enumerate}
    \item Answer: <ans> [answer text] <ans>
\end{enumerate}

\noindent Context Passage 1: [Context Passage 1] 

\noindent Context Passage 2: [Context Passage 2 if available]

\noindent Sentence: [Question to be evaluated] 

\noindent Response:

\subsection{NACo Details}\label{app:naco_c}
For each question, the Chain-of-Thought (CoT) QA prompt we provide to the LLM asks the model to output its question-answering process by steps, separated by newline characters. We post-process this formatted output to count the number of reasoning steps, referred to as the absolute complexity of the candidate question or $c_{\text{cand} \textunderscore \text{abs}}$.

To calculate the relative complexity of the candidate question with respect to the dataset, we first find the expected complexity associated with that dataset. Using a set of reference questions from the training set, we perform the same CoT QA process for each of these reference questions and obtain their absolute complexity. The expected complexity for the dataset is then the most common value (or mode) among the absolute complexities of the questions in this training sample, denoted as $c_{\text{expected}}$.

The final score regarding the complexity of the candidate question is the normalized value of the absolute difference between $c_{\text{cand} \textunderscore \text{abs}}$ and $c_{\text{expected}}$: $c_{\text{cand}} = 1- \frac{ | c_{\text{cand} \textunderscore \text{abs}} - c_{\text{expected}} |}{\max ( c_{\text{cand} \textunderscore \text{abs}} , c_{\text{expected}} ) }$. By using $\max ( c_{\text{cand} \textunderscore \text{abs}} , c_{\text{expected}} )$, we ensure the range of $c_{\text{cand}}$ is between $0$ and $1$. The final NACo score is then computed by taking a weighted sum of $n_{\text{cand}}$ (binary, $0$ or $1$), $a_{\text{cand}}$ (floating number between $0$ and $1$), and $c_{\text{cand}}$ (floating number between $0$ and $1$). We used a weight of $\frac{1}{3}$ for each criterion score in our experiments, ensuring NACo's range to be between $0$ and $1$. In short: $\text{NACo} =\frac{1}{3} n_{\text{cand}} +\frac{1}{3} a_{\text{cand}} + \frac{1}{3} c_{\text{cand}}$.

\subsection{Human Evaluation Details}

\subsubsection{Instructions}
\label{app:survey_instructions}

In this survey, you will be annotating 10 examples. For each example, you are given 2 passages that share some common information. A text span from one of the two passages will be bolded, italicized, and highlighted in blue. Your task is to rate 4 candidate questions on a scale of 0-2 for each of the following aspects:

\noindent \textbf{Fluency}: Does the question make at least one of the following errors: (1) grammar mistakes, (2) unclear objectives, or (3) not a question?
\begin{itemize}
    \setlength\itemsep{-0.5em}
    \item If the question does not make any errors, give a 2 for this criterion
    \item If the question makes 1 of the above errors, give a 1 for this criterion
    \item If the question makes at least 2 of the above errors, give a 0 for this criterion
\end{itemize}

\noindent \textbf{Answerability}: Try answering each question yourself. An acceptable question should be relevant to the context passages and has a reasoning path that leads to the given answer highlighted in blue.
\begin{itemize}
    \setlength\itemsep{-0.5em}
    \item If the answer to the candidate question is exactly the text highlighted in blue, give a 2 for this criterion
    \item If the answer to the candidate question contains some but not all parts of the text highlighted in blue, or contains all parts of the text highlighted in blue but with extra information, give a 1 for this criterion
    \item If the answer to the candidate question does not match the text highlighted in blue at all, give a 0 for this criterion.
\end{itemize}

\noindent \textbf{Complexity}: Try answering each question yourself. Does the question require reasoning over both passages? An acceptable question should use information from both passages, not just one.
\begin{itemize}
    \setlength\itemsep{-0.5em}
    \item If you need to read both passages to answer the question, give a 2 for this criterion
    \item If you need to read only one passage to answer the question, give a 1 for this criterion.
    \item If you do not need any of the passages to answer the question, give a 0 for this criterion.
\end{itemize}

\subsubsection{Human Evaluation Results}\label{app:human_eval_results}
\begin{table}[ht]
\centering
\begin{tabular}{lcccc}
\hline
\textbf{QG Competitor} & \textbf{Nat.} & \textbf{Ans.} & \textbf{Cmp.} & \textbf{Total} \\
& \textbf{[0,2]}       & \textbf{[0,2]}         & \textbf{[0,2]}       & \textbf{[0,6]}\\
\hline
BART-base & 1.10 & 0.74 & 0.97 & 2.80 \\
GPT-3.5   & 1.92 & 1.62 & 1.06 & 4.60 \\
RefQ      & 1.70 & 1.68 & 1.75 & 5.14 \\
AnnoQ     & 1.82 & 1.83 & 1.80 & 5.45 \\
\hline
\end{tabular}
\caption{Human Evaluation of QG Competitors on HotpotQA}
\label{tab:qg_competitors}
\end{table}

\subsection{TedEdQA Details}
\label{app:teded}

We collect 4246 multiple-choice questions from 1001 video lessons from TED-Ed\footnote{\url{https://ed.ted.com/}}. Each data point comprises the transcript of the video lesson it is based on, the question stem, and the correct answer.. After excluding questions with answers such as \textit{None of the above}, \textit{All of the above}, \textit{Both A and B}, etc., 3547 questions remain. We split the questions into three sets train, dev, and test, each with size of 3034, 259, and 254 respectively. We ensure that no questions from any set come from the same lecture as those in the other two sets.

From the test set, we select 43 questions (RefQ) derived from 12 video lessons for additional reference annotation. We follow similar procedures to the SQuAD and HotpotQA dataset that have annotators create two types of questions—one without clues and one with provided clues—based on a given context and answer. However, the context presented to annotators differs: to formulate a reference-qualifying question, we provide them with the URL of the original lesson, the full transcript, and a specific context extracted from the transcript that is relevant to the answer. This extraction is conducted as the entire video transcript can be too long, potentially complicating the fine-tuning of models like BART. We obtain this extracted context by having GPT3.5-turbo label it from the full transcript and the original question RefQ. Specifically, we prompt the model: ``\textit{Given a lecture content and a multiple-choice quiz question, please extract the most relevant and concise context from the content that is best for creating the provided multiple-choice question. Ensure the extracted excerpt contains all the necessary information for creating the given quiz question}''. This context is also used to fine-tune BART-base and to generate questions with GPT-3.5-turbo in a few-shot setting.

\subsection{Experiment Details}\label{app:exp_details}

Our \textbf{BART-base QG models} are initialized from checkpoint \verb|facebook/bart-base|, which has 139M parameters, and further finetuned on the specific QG dataset (SQuAD or HotpotQA). All models are implemented with Hugging Face Transformers 4.20. We add two special tokens: (1) \verb|<ans>| - used to highlight the answer span in the context input, and (2) \verb|<clue>| - used to highlight the clue words in the context input (for BART-clue-RefQ). The model is finetuned with a batch size of $128$, a learning rate of $1e-4$, a maximum input length of $512$, and a maximum output length of $32$. The best model is selected based on the lowest validation loss. 


\textbf{Implementations of existing metrics}: We use the implementation of Hugging Face \verb|evaluate|\footnote{\url{https://huggingface.co/docs/evaluate/en/index}} package for BLEU (\verb|bleu|), ROUGE (\verb|rouge|), BLEURT (\verb|bleurt|), BERTScore (\verb|bertscore|), and RQUGE (\verb|rquge|). We use the code released by the original papers to obtain implementation of QAScore\footnote{\url{https://github.com/TianboJi/QAScore/tree/main}} and Q-BLEU\footnote{\url{https://github.com/PrekshaNema25/Answerability-Metric}}. 

For Div-Ref, which proposes diversifying references using LLM to improve reference-based metrics' alignment with human judgement, we use the same model settings as the authors \cite{tang2023metrics}. Specifically, we use GPT3.5-turbo with temperature set to 1 and top\_p set to 0.9. We use 9/10 instructions proposed by \citealt{tang2023metrics} to generate 9 new references from the original reference RefQ. (We did not use the remaining instruction because it asks the model to reorder sentences in a paragraph, while our text is only a question in the form of \underline{one} sentence). When calculating the reference-based metric score across multiple references, we used the maximum aggregation.

\textbf{LLM Details}: We test 5 different LLMs for NACo. We interact with GPT3.5 (\verb|gpt3.5-turbo|), GPT4o (\verb|gpt-4o|)\footnote{\url{https://platform.openai.com/docs/overview}}, Claude-Haiku (\verb|claude-3-haiku-20240307|)\footnote{\url{https://www.anthropic.com/api}}, and Mixtral-8x7B (\verb|open-mixtral-8x7b|)\footnote{\url{https://docs.mistral.ai/api/}} through their official APIs. For Llama3-8B, we download the model via their Huggingface repository (\verb|meta-llama/Meta-Llama-3-8B-Instruct|) and deploy it locally.
All experiments with LLMs are used with their default hyperparameters. In our CoT-QA experiments on SQuAD, given the larger sample size of 750 and cost constraints, we conducted a single run. For our CoT-QA experiments on HotpotQA, we carried out 3 runs on the 50-examples sample and report the average scores from the three responses.

\subsection{Error Analysis}\label{app:error_analysis}

\begin{figure}[t]
  \centering
  \includegraphics[width=0.9\linewidth]{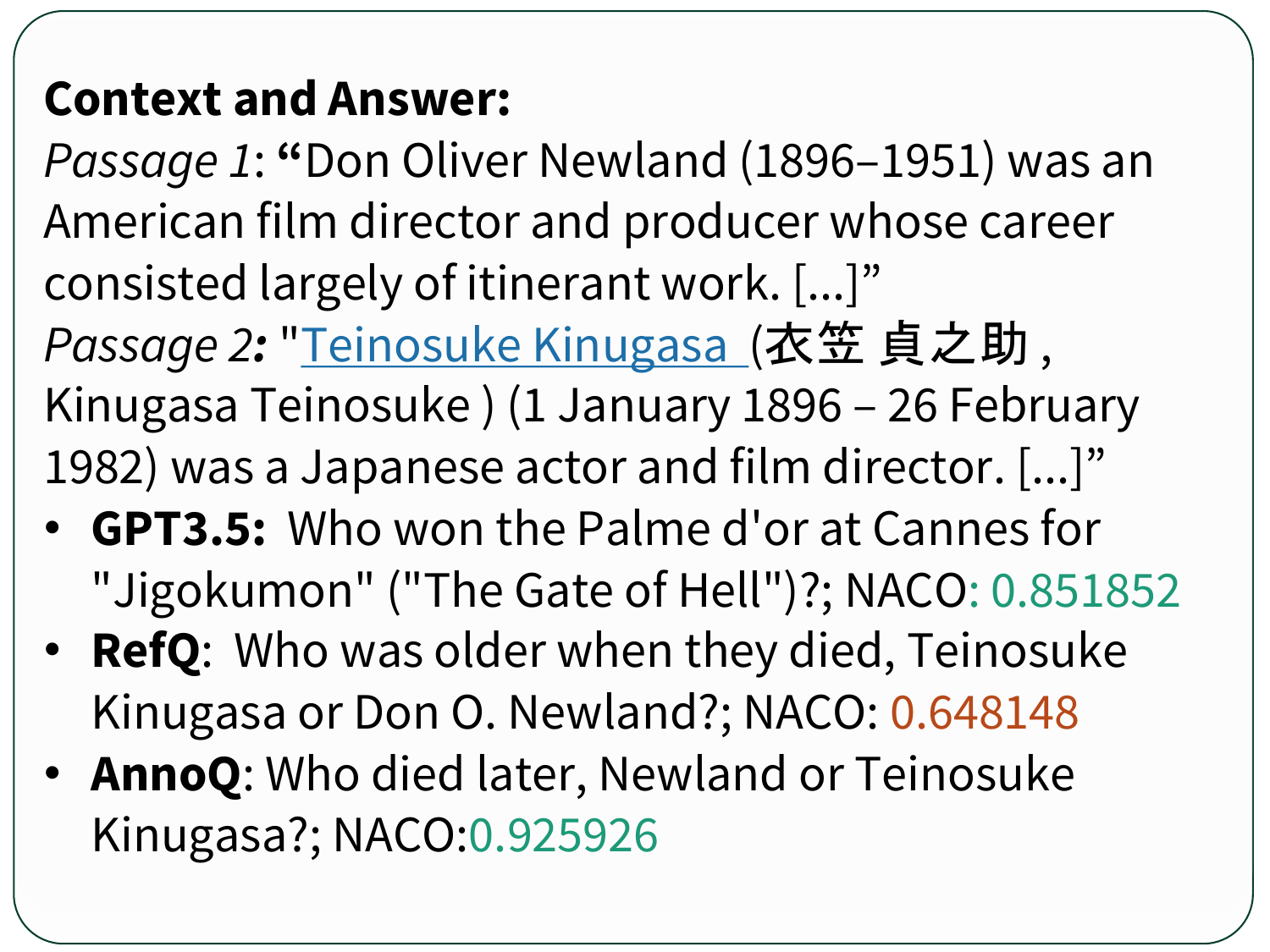}
  \caption{NACo Error Analysis: Reliance on QA model capacity.}
  \label{fig:naco_error}
\end{figure}

We have noted that one of the limitations of NACo is the dependency of the QA models' performance. To further elaborate it, we provide the a case study in Fig. \ref{fig:naco_error}.
The case study involves two context passages and the answer `\textit{Teinosuke Kinugasa}.' We examine three candidate questions: a \textbf{GPT-3.5-generated} question (intended for 2-hop but resulting in 1-hop), the original HotpotQA reference (\textbf{RefQ}, 2-hop), and our newly collected reference (\textbf{AnnoQ-clue-RefQ}, 2-hop). The CoT-QA model employed by NACo (GPT3.5) correctly identifies 'Teinosuke Kinugasa' for both the GPT-3.5-generated question and AnnoQ-clue-RefQ but fails to do so for RefQ, responding with `\textit{Not enough information provided to answer the question}.' 

This failure with RefQ is attributed to its requirement for mathematical reasoning (subtracting birth year from death year), a task GPT-3.5 struggles with. Accordingly, NACo assigns the highest score to AnnoQ-clue-RefQ, fulfilling all three requirements, while penalizing the GPT-3.5 question for its simplicity and RefQ most severely (Answerability F1 score = 0) due to the mismatch between the CoT-QA answer and the provided answer.

\subsection{Additional results for reference-based metrics}

Tbl. \ref{result_squad_refq_full} and \ref{result_hotpot_refq}  provides a more detailed version of \ref{result_squad_refq} and \ref{result_hotpot_refq_small}.

Tbl. \ref{result_teded_refq} illustrates the application of our study that disproves reference-based metrics and the proposed metric, NACo, in an educational setting using the TedEd-QA dataset. It can be seen that our observations regarding the failure of reference-based metrics and the effectiveness of NACo also holds for this dataset. Specifically, Refq, AnnoQ, and AnnoQ-clue-RefQ have significant gap when reference-based metrics are used to score them. NACo is able to score all these three human-validated candidates with similar scores and no worse than any machine-generated candidates.

\begin{table*}[t]
    \begin{center}
    \resizebox{0.8\textwidth}{!}{%
    \begin{tabular}{lrrrrr|rrr}
    \hline
    \noalign{\smallskip}
    \multirow{2}{*}{QG Competitor} & \multicolumn{5}{c|}{Ref-based metrics} & \multicolumn{3}{c}{Ref-free metrics} \\
         & B & B-RT & R-L & BSc & Q-B & QA-S & R-Q & 
         NACo \\ 
         \noalign{\smallskip}
         \hline
         \noalign{\smallskip}
        \textbf{LM-generated} &  &  &  &  &  &  &  &   \\ 
        BART-base & 19.53 & -0.28 & 44.79 & 92.13 & 36.94 & \textbf{-0.37} & 4.62 & 
        73.30\\ 
        GPT-3.5 (few-shot) & 18.06 & -0.23 & 43.58 & 92.18 & 36.48 & \textbf{-0.37 }& 4.56 & 
        73.67\\ 
        BART-clue-RefQ & \textbf{31.91} & \textbf{0.07} & \textbf{59.92} & \textbf{94.37} & \textbf{52.33} & -0.38 & 4.56 & 
        69.97 \\ 
        \noalign{\smallskip}
        \hline
        \noalign{\smallskip}
        \textbf{Human-validated} &  &  &  &  &  &  &  &   \\ 
        RefQ & 100.00 & 1.00 & 100.00 & 100.00 & 100.00 & -0.38 & \textbf{4.89} & 
        \textbf{75.09} \\ 
        AnnoQ & 12.78 & -0.31 & 37.83 & 91.52 & 31.32 & \textbf{-0.37 }& 4.71 & 
        74.01 \\ 
        AnnoQ-clue-RefQ & \underline{27.43} & \underline{0.04} & \underline{53.62} & \underline{93.85} & \underline{46.89} & -0.38 & \underline{4.76} & 
        \underline{74.21}\\ 
        \noalign{\smallskip} \hline
    \end{tabular}}
    \caption{\label{result_squad_refq_full}\textbf{SQuAD} - Performance of different QG methods on NACo and other existing metrics. The evaluation uses original SQuAD questions (RefQ) as references, with GPT-3.5 as the underlying QA system for NACo. The highest and second-highest scores (not including references for reference-based metrics) are highlighted with bold and underline markers, respectively.}
    \end{center}
\end{table*}

\begin{table*}[t]
    \begin{center}
    \resizebox{0.85\textwidth}{!}{%
     \begin{tabular}{lrrrrr|rrr|r}
    \hline
    \noalign{\smallskip}
     \multirow{2}{*}{QG Competitor} & \multicolumn{5}{c|}{Ref-based metrics} & \multicolumn{3}{c|}{Ref-free metrics} & Human  \\
         & B & B-RT & R-L & BSc & Q-B & QA-S & R-Q & 
         NACo  \\ \noalign{\smallskip} \hline \noalign{\smallskip}
                 \textbf{LM-generated} &  &  &  &  &  &  &  & &   \\ 

        BART-base & 14.57 & -0.75 & 34.87 & 87.83& 28.58 & -0.28 & 2.90 & 
        42.26 & 2.80\\ 
        GPT-3.5 (zero-shot) & 9.46 & -0.92 & 26.95 & 87.48 & 16.87 & -0.28 & 4.18 & 
        74.99 & 4.60\\ 
        GPT-3.5 (few-shot) & 9.55 & -0.86 & 27.48 & 87.71 & 17.33 & -0.27 & 4.33 & 
        77.41 &  - \\ 
        BART-clue-RefQ & \textbf{34.29} & \underline{-0.07} & \textbf{61.40} & \underline{92.86} & \textbf{53.94} & -0.28 & 3.51 & 
        62.40 & - \\ 
        \noalign{\smallskip} \hline \noalign{\smallskip}
        \textbf{Human-validated} &  &  &  &  &  &  &  &  \\ 
        RefQ & 100.00 & 1.00 & 100.00 & 100.00 & 100.00 & -0.28 & 4.12 & 
        75.32 & 5.14\\ 
        AnnoQ & 14.80 & -0.59 & 34.21 & 89.65 & 28.77 & \textbf{-0.27} & \underline{4.21} & 
        \textbf{84.97} & 5.45 \\ 
        AnnoQ-clue-RefQ & \underline{32.56} & \textbf{-0.05} & \underline{53.21} & \textbf{93.12} & \underline{52.85} & \textbf{-0.27} & \textbf{4.26} &  
        \underline{79.16} & -  \\ 
        \noalign{\smallskip} \hline
    \end{tabular}}
    \caption{\label{result_hotpot_refq} \textbf{HotpotQA} -  Performance of different QG methods on NACo and other existing metrics. The evaluation uses original HotpotQA questions (RefQ) as references, with GPT-3.5 as the underlying QA system for NACo. The highest and second-highest scores (not including references for reference-based metrics) are highlighted with bold and underline markers, respectively.}
    \end{center}
\end{table*}

\begin{table*}[t]
    \begin{center}
    \resizebox{\textwidth}{!}{%
     \begin{tabular}{lrrrrr|rr|rrrr}
    \hline
    \noalign{\smallskip}
     \multirow{2}{*}{QG Competitor} & \multicolumn{5}{c|}{Ref-based metrics} & \multicolumn{2}{c|}{Ref-free metrics} & \multicolumn{4}{c}{Our metric}  \\
         & B & B-RT & R-L & BSc & Q-B & QA-S & R-Q  & NACo  \\ \noalign{\smallskip} \hline \noalign{\smallskip}
                 \textbf{LM-generated} &  &  &  &  &  &  &  &  \\ 

        BART-base & 13.71 & \underline{0.16} & 32.42 & 89.19 & - & -\textbf{0.15 }& 3.74 & 79.29 \\ 
        GPT-3.5 (few-shot) & 9.23 & -0.28 & 28.77 & 88.6 & - & -0.16 & 3.89 &  79.78 \\ 
        BART-clue-RefQ & \underline{13.63} & 0.01 & \underline{40.66} & \underline{90.09} & - & \textbf{-0.15} & 3.33  & 75.29 \\ 
        \noalign{\smallskip} \hline \noalign{\smallskip}
        \textbf{Human-validated} &  &  &  &  &  &  & &  \\ 
        RefQ & 100.00 & 1.00 & 100.00 & 100.00 & - & -0.16 & \underline{3.99} & 82.85 \\ 
        AnnoQ & 14.78 & -0.43 & 34.16 & 89.61 & - & -0.16 & 3.98 & \textbf{84.67} \\ 
        AnnoQ-clue-RefQ & \textbf{26.4} & \textbf{0.59}  & \textbf{50.22} & \textbf{92.1} & - & -0.17 & \textbf{4.23}  & \underline{83.00} \\ 
        \noalign{\smallskip} \hline
    \end{tabular}}
    \caption{\label{result_teded_refq} \textbf{TedEdQA} -  Performance of different QG methods using NACo and other existing metrics. The evaluation uses original TedEd questions (RefQ) as references, with GPT-3.5 as the underlying QA system for NACo. Some questions in this dataset are in the fill-in-the-blank form and do not contain question words like \textit{what}, \textit{where}, \textit{etc}. which Q-BLEU heavily relies on for scoring \cite{nema-khapra-2018-towards}; thus, we do not report this metric for this dataset. The highest and second-highest scores (not including references for reference-based metrics) are highlighted with bold and underline markers, respectively.}
    \end{center}
\end{table*}
\end{document}